# MECHANISM FOR EMBOSSING BRAILLE CHARACTERS ON PAPER: CONCEPTUAL DESIGN

Jonnatan Arroyo and Ramiro Velázquez
*Mechatronics and Control Systems Lab, Universidad Panamericana,
Aguascalientes, Mexico*

## INTRODUCTION

This paper presents the conceptual design of a low-cost simple printer head for Braille embossers. Such device consists of a set of three rotary cam-follower mechanisms that, upon actuation, produce deformation on paper. The set of cam-followers is actuated by a single servomotor which rotation determines which cam-follower strikes the paper. Braille characters are quickly embossed by column using the proposed system. The aim of this research is to provide new actuation ideas for making Braille embossers more affordable.

## BACKGROUND

There are several ways to transmit text information to the blind. For text displayed on a PC screen, voice synthesizers, screen magnifiers, and Braille output terminals are popular solutions. Voice synthesizers practically read the computer screen; screen magnifiers enable on-screen magnification for those with low-vision, and Braille output terminals are plugged to the computer so that information on the screen is displayed in Braille (Brabyn, 2007; Velazquez, 2008).

The easiest way -one can think of- to transmit text information to the blind is perhaps by audio books. Entire text books are recorded in the form of speech and reproduced by standard music players (wearable headsets or desktop players).

Since 1950, audio books have benefited the blind population by offering a simple, low-cost, non-Braille reading option. Thousands of titles are available, and with today's technology they can be downloaded for example, from the National Library Service for the Blind and Physically Handicapped (NLS, 2011). Even though the undisputable advantages of audio books, two issues have to be considered (Velazquez, 2010):

- Availability. Even though there are thousands of titles, not all books are systematically converted to audio books and there is a significant delay with new books. Months, years (or never) could take to have a new book available in audio form.

- Audio books should not be considered as the reading solution for blind people. It is true that they fit perfectly for the elderly and non-Braille readers. However as healthy-sighted do prefer to read a book instead of hearing it, why do we assume that a blind person prefers to listen to a book instead of reading it? For the blind, reading a book heightens the self-esteem and independence. Moreover, it trains their orthography which is in most cases quite bad.

Another way to transmit text information is by tactile printing. Both text and graphics can be embossed on paper using impact printers, which hit the paper surface with sufficient force to create a deformation that can be read with the fingertips. Their main inconvenient is their cost: they can cost roughly anywhere from 2,000 to 80,000 USD (NFB, 2011). In consequence, tactile embossers are rarely privately owned. They can be found in some libraries, universities, and special education centers. Cost drawbacks remain: printing a single page in these places is 5 times more expensive than printing a page with traditional printers. Moreover, industrial fabrication of tactile print books is nowadays poor in comparison to that of audio books.

One of the key features of tactile embossers is the printer head that strikes the paper to form the Braille dots or tactile graphics. Printer heads of commercial embossers are mainly based on high force performance solenoids called hammer solenoids. Unfortunately, this technology has become too expensive over the years.

With the aim of envisioning more affordable Braille embossers, we propose a low-cost simple printer head system capable of embossing Braille code characters on paper. This system is based on traditional mechanisms such as the cam-follower and low-cost commercially available actuators such as the servomotor. In this paper, we introduce the conceptual design and we briefly discuss its ongoing implementation.

## SYSTEM OVERVIEW

The prototype shown in fig. 1(a) is the conceptual design of our printer head for Braille embossers.

It is a compact device (73 x 39.5 x 39 mm) that complies with Braille standards (2.54 mm pin interspacing) and, contrary to commercial embossers, it is intended to be fixed to the printer's frame while the paper progressively moves under it. As the paper moves, its three pins strike sequentially to form the Braille dots on one side of the paper. Braille dots are embossed by column, so a Braille character would require three sequences: one strike, then paper displacement, and another strike.

Paper displacement is to be carried out by two step motors: one for moving the paper from side to side (in order to print lines) and one for moving the paper forward (in order to change to the next line).

The structure of the printer head is shown in fig. 1(b). A single servomotor transfers rotational motion to a shaft through a gear train. Three cam-follower mechanisms circumferentially arranged at 120° are located around this shaft and are actuated according to the rotation performed by the servomotor. Two optical encoders provide feedback on the shaft's angular position to ensure the activation of the desired cam-follower.

Fig. 2 details the operation principle of the proposed printer head. Miniature cam-follower mechanisms are quickly actuated to emboss 6-dot Braille characters on paper.

This actuation approach offers some interesting advantages over the existing ones:

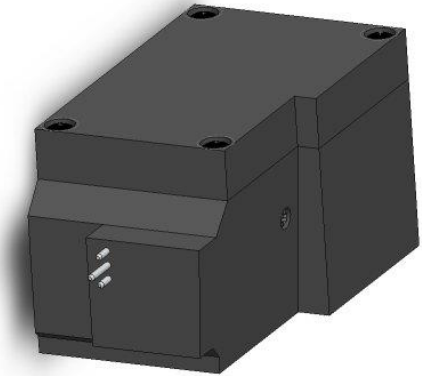

(a)

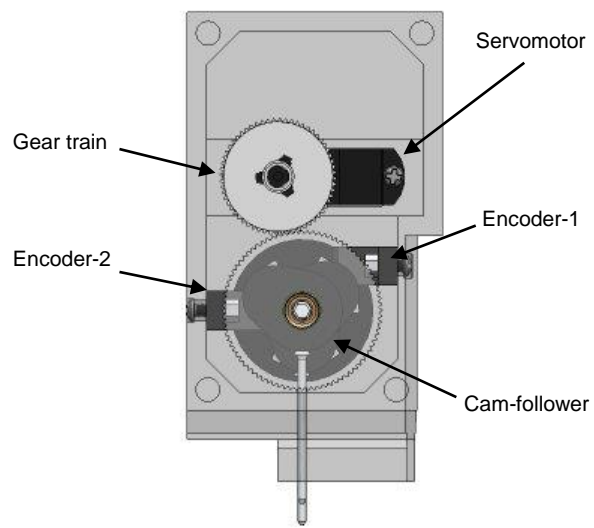

(b)

Figure 1: Conceptual design of the set of servo-actuated cam-follower mechanisms for embossing Braille characters on paper.

it is simple, it involves only one actuator to control, it offers noiseless operation, it is low-cost, and fast/easy to implement. However, due to the use of a single actuator, the expected printing speed is suitable for home use only (industrial Braille embossers can output up to 800 Braille characters per second using up to 800 actuators).

Implementation of this design is currently in process. The actuator selected is a high-torque servomotor (3.5 kg-cm) that ensures the correct striking force to emboss Braille characters and a suitable speed (0.11 s/60°) for this application.

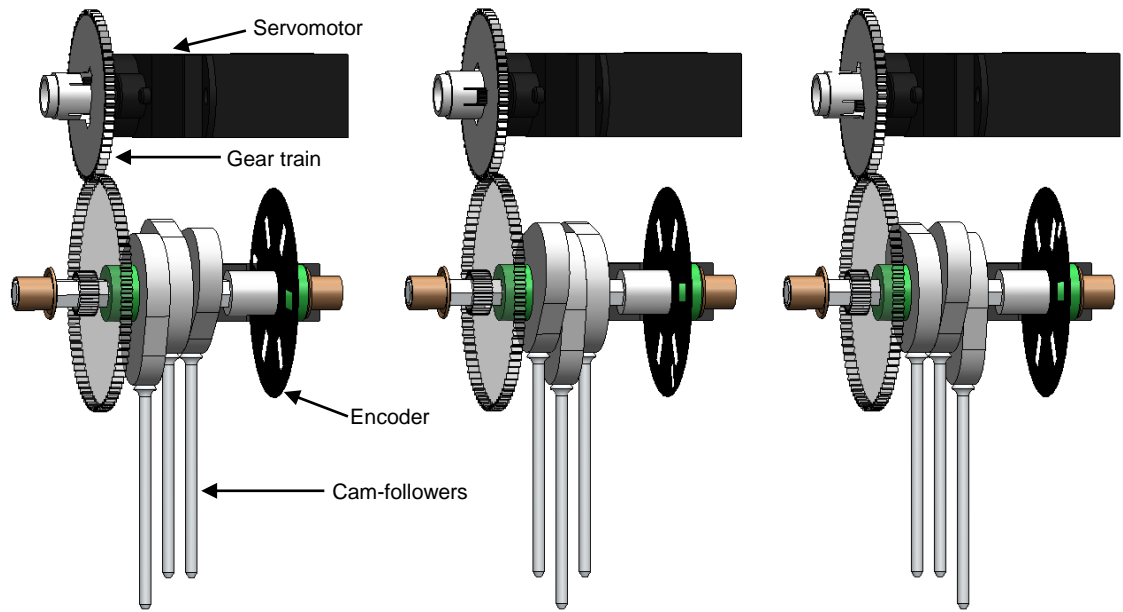

Figure 2: Detail of the embossing system. Three cam-follower mechanisms are arranged at 120° intervals around the shaft and their actuation is determined by the rotation of the servomotor.

Moreover, its dimensions and mass (29.5 x 11.6 x 30.2 mm, 17.5 g) are in accordance to the compact printer head design.

Gear train and cam-followers are being first rapid-prototyped in thermoplastic material. A second stage would require their implementation in metal. All parts are to be integrated in a hard plastic enclosure of compact dimensions (73 x 39.5 x 39 mm). The complete printer head prototype is expected not to exceed 40 USD.

## CONCLUSION

Developing new Braille technologies is important for improving the accessibility of information for blind people. We focus our attention on actuators that can be used for displaying text information in Braille. In particular, this paper introduced the design of a novel printer head system that embosses Braille characters on paper.

Such system consists of three servo-actuated rotary cam-follower mechanisms that strike the paper to form Braille dots. The concept is simple, noiseless, low-cost, and is the base for the development of a truly affordable Braille embosser prototype intended for home use.

Current work focuses on the implementation and integration of its components.